\documentclass[a4paper,UKenglish]{tgdk}
\usepackage{fontawesome5}

\title{Knowledge Engineering using Large Language Models} 

\author{Bradley P. Allen}{University of Amsterdam, Amsterdam, NL}{b.p.allen@uva.nl}{https://orcid.org/0000-0003-0216-3930}{}

\author{Lise Stork}{Vrije Universiteit Amsterdam, Amsterdam, NL}{l.stork@vu.nl}{https://orcid.org/0000-0002-2146-4803}{}

\author{Paul Groth}{University of Amsterdam, Amsterdam, NL}{p.t.groth@uva.nl}{https://orcid.org/0000-0003-0183-6910}{}

\authorrunning{B.\,P. Allen, L. Stork, P. Groth}

\Copyright{Bradley P. Allen, Lise Stork, Paul Groth}

\subjclass{}


\DOI{10.1234/0000000.00000000}
\DateSubmission{2023-07-14} 
\nolinenumbers

\begin{document}

\maketitle

\begin{abstract}
Knowledge engineering is a discipline that focuses on the creation and maintenance of processes that generate and apply knowledge. Traditionally, knowledge engineering approaches have focused on knowledge expressed in formal languages. The emergence of large language models and their capabilities to effectively work with natural language, in its broadest sense,
raises questions about the foundations and practice of knowledge engineering. Here, we outline the potential role of LLMs in knowledge engineering, identifying two central directions: 1) creating hybrid neuro-symbolic knowledge systems; and 2) enabling knowledge engineering in natural language. Additionally, we formulate key open research questions to tackle these directions.
\end{abstract}

\section{Introduction}
Knowledge engineering (KE) is a discipline concerned with the development and maintenance of automated processes that generate and apply knowledge \cite{allen2023identifying,studer1998knowledge}. Knowledge engineering rose to prominence in the nineteen-seventies, when Edward Feigenbaum and others became convinced that automating knowledge production through the application of research into artificial intelligence required a domain-specific focus \cite{feigenbaum1977art}. The period from the mid-nineteen-seventies into the nineteen-eighties saw the knowledge engineering of rule-based expert systems for the purposes of the automation of decision making in business enterprise settings. By the early nineteen-nineties, however, it became clear that the expert systems approach, given its dependence on manual knowledge acquisition and rule-based representation of knowledge by highly skilled knowledge engineers, resulted in systems that were expensive to maintain and difficult to adapt to changing requirements or application contexts. Feigenbaum argued that, to be successful, future knowledge-based systems would need to be scalable, globally distributed, and interoperable \cite{feigenbaum1992personal}. 

The establishment of the World Wide Web and the emergence of Web architectural principles in the mid-nineteen-nineties provided a means to address these requirements. Tim Berners-Lee argued for a "Web of Data" based on linked data principles, standard ontologies, and data sharing protocols that established open standards for knowledge representation and delivery on and across the Web \cite{berners2001semantic}. The subsequent twenty years witnessed the development of a globally federated open linked data "cloud" \cite{bizer2008linked}, the refinement of techniques for ontology engineering \cite{kendall2019ontology}, and methodologies for the development of knowledge-based systems \cite{schreiber2000knowledge}. During the same period, increasing use of machine learning and natural language processing techniques led to new means of knowledge production through the automated extraction of knowledge from natural language documents and structured data sources \cite{collobert2011natural,nasar2021named}. Internet-based businesses in particular found value in using such technologies to improve access to and discovery of Web content and data \cite{hendler2020semantic}. A consensus emerged around the use of knowledge graphs as the main approach to knowledge representation in the practice of knowledge engineering in both commercial and research arenas, providing easier reuse of knowledge across different tasks and a better developer experience for knowledge engineers \cite{hogan2021knowledge}.

More recently, the increase in the availability of graphical processing hardware for fast matrix arithmetic, and the exploitation of such hardware to drive concurrent innovations in neural network architectures at heretofore unseen scales \cite{wang2017origin}, has led to a new set of possibilities for the production of knowledge using large language models (LLMs). LLMs are probabilistic models of natural language, trained on very large corpora of content, principally acquired from the Web. Similiar to previous approaches to language modelling , given a sequence of tokens LLMs predict a probable next sequence of tokens based on a learned probability distribution of such sequences. However, presumably due to the vast amount of content processed in learning and the large size and architecture of the neural networks involved, LLMs exhibit remarkable capabilities for natural language processing that far exceed earlier approaches \cite{mahowald2023dissociating}. 

These capabilities include the ability to do zero- or few-shot learning across domains \cite{brown2020language}, to generalize across tasks, including the ability to perform domain-independent question answering integrating large amounts of world knowledge \cite{radford2021learning}, to generate text passages at human levels of fluency and coherence \cite{devlin2018bert,sutskever2014sequence}, to deal gracefully with ambiguity and long-range dependencies in natural language \cite{vaswani2017attention}, and to reduce or even eliminate the need for manual feature engineering \cite{tenney2019bert}. LLMs also exhibit the ability to generate and interpret structured and semi-structured information, including programming language code \cite{austin2021program,vaithilingam2022expectation}, tables \cite{hulsebos2019sherlock,korini2023column}, and RDF metadata \cite{wang2017origin,lorandi2023data,axelsson2023using}. The generalization of language models (termed "foundation models" by some) to other modalities including images and audio have led to similarly significant advances in image understanding \cite{caron2021emerging,yu2022coca}, image generation \cite{goodfellow2020generative,ramesh2022hierarchical,rombach2022high}, speech recognition, and text-to-speech generation \cite{radford2023robust,wang2023neural}. Such capabilities have prompted a significant amount of research and development activity demonstrating potential applications of LLMs \cite{nlpPLMSurvey2023,schick2023toolformer,kung2023performance}. However, the means of incorporating LLMs into structured, controllable, and repeatable approaches to developing and fielding such applications in production use are only just beginning to be considered in detail \cite{pan2023unifying}. 

This paper engages with the question of how LLMs can be effectively employed in the context of knowledge engineering. We start by examining the different forms that knowledge can take, both as inputs for constructing knowledge systems and as outputs of such systems. We argue that the distinction between knowledge expressed in natural language (or other evolved, naturally occurring modalities such as images or video) and knowledge expressed in formal languages (for example, as knowledge graphs or rules), sheds light how LLMs can be brought to bear on the development of knowledge systems.

Based on this perspective, we then describe two potential paths forward. One approach involves treating LLMs as components within hybrid neuro-symbolic knowledge systems. The other approach treats LLMs and prompt engineering \cite{liu2023pre} as a standalone approach to knowledge engineering\footnote{As defined by \cite{liu2023pre}, prompt engineering is finding the most appropriate prompt or input text to an LLM to have it solve a given task.}, using natural language as the primary representation of knowledge. We then enumerate a set of open research problems in the exploration of these paths. These problems aim to determine the feasibility of and potential approaches to using LLMs with existing KE methodologies, as well as the development of new KE methodologies centered around LLMs and prompt engineering.

\section{Forms of knowledge and their engineering}

In the history of the computational investigation of knowledge engineering, knowledge has been often treated primarily as symbolic expressions. However, as \cite{groth2023forms} noted, knowledge is actually encoded in a variety of media and forms, most notably in natural language (e.g. English) but also in images, video, or even spreadsheets. This fact becomes even more apparent when looking at institutional knowledge practices that have developed over centuries, for example, in the sciences or archives \cite{hjorland2008knowledge}. We now illustrate this point by describing the many ways in which knowledge manifests itself in the context of biodiversity informatics.

\subsection{The multimodal richness of knowledge: an example from biodiversity sciences}\label{sec:bio}

 The ultimate goal of biodiversity science is to understand species evolution, variation, and distribution, but finds applications in a variety of other fields such as climate science and policy. At its heart is the collection and observation of organisms, providing evidence for deductions about the natural world~\cite{macgregor2018naturalists}. Such knowledge is inherently multimodal in nature, most commonly appearing in the form of images, physical objects, tree structures and sequences, i.e., molecular data. 

 Historically, organism sightings have been carefully logged in handwritten field diaries to describe species behaviour and environmental conditions. Detailed drawings and later photographs were made to capture colour, organs and other knowledge about an organism's traits used for identification, which is best conveyed visually but which is challenging to preserve in natural specimens. These manuscripts are housed, together with the physical zoological specimens and herbaria which they describe, in museums and collection facilities across the world. Both the multimodal nature of these knowledge sources as well as their distributed nature hamper knowledge integration and synthesis. 
 
\begin{figure}[ht]%
\begin{minipage}{\textwidth}
    \centering    {\includegraphics[width=1\linewidth]{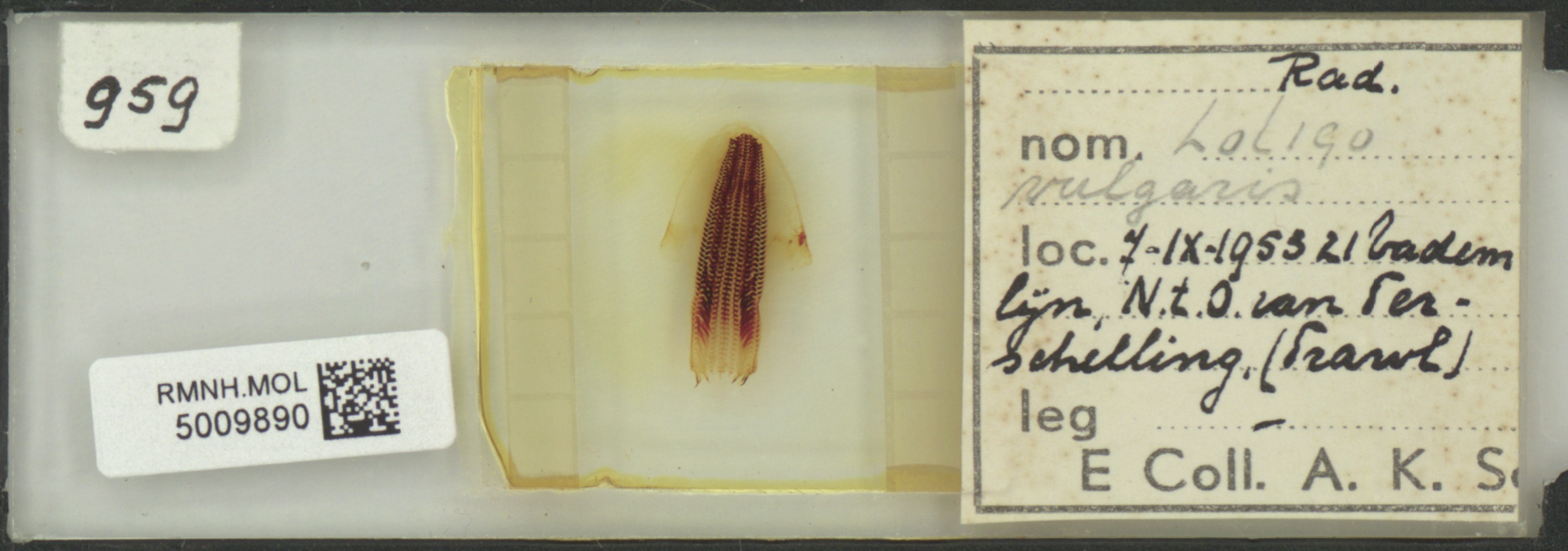}}
    \caption[A specimen of the \textit{Loligo vulgaris Lamarck, 1798} species from the \textit{Naturalis--Zoology and Geology} catalogues.]{A specimen of the \textit{Loligo vulgaris Lamarck, 1798} species from the \textit{Naturalis--Zoology and Geology} catalogues.\footnote{\url{https://bioportal.naturalis.nl/nl/specimen/RMNH.MOL.5009890}} Images free of known restrictions under copyright law (Public Domain Mark 1.0).}
    \label{fig:spec}%
    \end{minipage}
\end{figure}

 Metadata describes the specimen's provenance: where specimens were found, who found them, and provides an attempt at identifying the type of organism (such as the preserved squid specimen shown in Figure \ref{fig:spec}). Such knowledge is paramount, as it allows researchers to understand resources within the context in which they were produced, enabling researchers to carry out ecological studies such as distribution modeling over time. 

 For a systematic comparison of the multitude of resources available, the biodiversity sciences have had a long-standing tradition of developing information standards~\cite{muller2017names}.  From Linnaeus' Systema naturae mid 18th century as well as his formal introduction of zoological nomenclature, taxonomists have started categorizing natural specimens according to tree-like hierarchical structures. The process is challenging, given that biologist up until this day do not have a full picture of all living organisms on earth, and incomplete, naturally evolved and fuzzy knowledge is not easily systematized.  


The development of digital methods has opened up new pathways for comparison and analysis. Gene sequencing technology has lead biologist to the genetic comparison of species, by the calculation of ancestry and construction of evolutionary tree structures in the study of phylogeny~\cite{kapli2020phylogenetic}. More importantly, digital methods allowed the transfer of analog resources, such as specimen collection scans~\cite{blagoderov2012no} and metadata, to the digital world. Such techniques have furthered formalisation and thereby interoperability of collected data through the use of Web standards, such as globally unique identifiers for species names~\cite{page2008biodiversity} as well as shared vocabularies for data integration across collections~\cite{baskauf2017tdwg}. The Global Biodiversity Information Facility (GBIF) and their data integration toolkit serves as a great example of such integration efforts~\cite{telenius2011biodiversity,robertson2014gbif}. 
Currently, there is a large emphasis on linking up disparate digital resources in the creation of an interconnected network of digital collection objects on the Web, linked up with relevant ecological, environmental and other related data in support of machine actionability (i.e., the ability of computational systems to find, access, interoperate, and reuse data with minimal intervention) for an array of interdisciplinary tasks such as fact-based decision-making and forecasting~\cite{hardisty2022digital}.Using data standards for describing and reasoning over collection data can aid researchers counter unwanted biases via transparency. However, making data comply with data standards can also lead to oversimplification or reinterpretation~\cite{ortolja2022encoding}. 

Machine learning and knowledge engineering strategies can help to (semi-)automatically extract and structure biodiversity knowledge according \cite{vanErpNHthesis2010,stork2021knowledge}, for instance using state-of-the-art computer vision or natural language processing techniques as well as crowd-sourcing platforms for the annotation of field diaries and other collection objects with formal language~\cite{stork2019semantic,dijkshoorn2013personalized}. Nevertheless, a bottleneck in the digitization of collections and their use for machine actionability is the amount of work and domain expertise required for the formalisation of such knowledge, and the extraction from unstructured texts, images and video's. Historical resources, i.e. handwritten texts, pose an additional challenge, as they are exceptionally challenging to interpret within the current scientific paradigm~\cite{weber2018towards}. 


The variety and usefulness of different forms of knowledge both natural and formal and the challenges they pose is not limited to the biodiversity domain as described above. We see the same diversity happening in law \cite{rodriguez2020lynx}, medicine \cite{bonner2022review,annurev-biodatasci-010820-091627} and even self-driving vehicles \cite{badue2021self}. To summarize: 
\begin{itemize}

\item domain knowledge is often best represented in a variety of modalities, i.e., images, taxonomies, or free text, each modality with its own data structure and characteristics which should be preserved, and no easy 
way of integrating, interfacing with or reasoning over multimodal knowledge in a federated way exists;
\item provenance of data is paramount in understanding knowledge within the context in which it was produced; 
\item fuzzy, incomplete, or complex knowledge is not easily systematized; 
\item using data standards for describing and reasoning over collection data can aid researchers counter unwanted biases via transparency; 
\item making data comply with data standards can lead to oversimplification or reinterpretation; 
\item the production of structured domain knowledge, for instance from images or free text, requires domain expertise, and is therefore labour intensive and costly; 
\item knowledge evolves, and knowledge-based systems are required to deal with updates in their knowledge bases. 
\end{itemize}

\subsection{KE as the transformation of knowledge expressed in natural language into knowledge expressed in a formal language}

This sort of rich and complex array of modalities for the representation of knowledge has traditionally posed a challenge to knowledge engineers \cite{FEIGENBAUM1984}. Much of the literature on knowledge engineering methodology has focused on the ways in which knowledge in these naturally-occurring forms can be recast into a structured symbolic representation, e.g., using methods of knowledge elicitation from subject matter experts \cite{shadbolt2015knowledge}, for instance by the formulation of competency questions for analysing application ontologies~\cite{bezerra2013evaluating}. One way to think about this is as the process of expressing knowledge presented in a natural, humanly evolved language in a formally-defined language. This notion of the transformation of natural language into a formal language as a means of enabling effective reasoning has a deep history rooted in methodologies developed by analytical philosophers of the early twentieth century \cite{carus2007carnap, novaes2012formal}, but dating even further back to Liebniz's \textit{lingua rationalis} \cite{gabbay2004rise} and the thought of Ram{\'o}n Lull \cite{glymour1998ramon}. Catarina Dutilh Novaes \cite{novaes2012formal} has argued that formal languages enable reasoning that is less skewed by bias and held beliefs, an effect achieved through \textit{de-semantification}, i.e., the process of replacing terms in a natural language with symbols that can be manipulated without interpretation using a system of rules of transformation. Coupled with sensorimotor manipulation of symbols in a notational system, people can reason in a manner that outstrips their abilities unaided by such a technology. 

While Dutilh Novaes' analysis focuses on this idea of formal languages as a cognitive tool used by humans directly, e.g. through the manipulation of a system of notation using paper and pencil, she notes that this manipulation of symbols is the route to the mechanization of reasoning through computation. When externally manifested as a function executed by a machine through either interpretation by an inference engine, or through compilation into a machine-level language, this approach of formalization yields the benefits of reliability, greater speed and efficiency in reasoning. 

This idea captures precisely the essence of the practice of knowledge engineering: Starting from sources of knowledge expressed in natural language and other modalities of human expression, through the process of formalization \cite{kendall2019ontology,suarez2012introduction}, knowledge engineers create computational artifacts embodying this knowledge. These computational artifacts then enable us to reason using this knowledge in a predictable, efficient, and repeatable fashion. This is done either by proxy through the action of autonomous agents, or in the context of human-mediated decision-making processes. 

\subsection{LLMs as a general-purpose technology for transforming natural language into formal language}

Until recently, there have been two ways in which this sort of formalization could be performed: through the manual authoring of symbolic/logical representations, e.g., as in the traditional notion of expert systems \cite{feigenbaum1992personal}, or through the use of machine learning and natural language processing to extract such representations automatically from natural language text \cite{MartinezRodriguez2020}. But what has become evident with the emergence of LLMs, with their capabilities for language learning and processing, is that they provide a new and powerful type of general purpose tool for mapping between natural language\footnote{Again, we note that natural language should be read to include all modalities. Hence, the term ``foundation model''\cite{foundationmodel2021} to refer to LLMs.} and formal language, as well as other modalities. LLMs have shown state-of-the-art performance on challenging NLP tasks such as relation extraction~\cite{alt2019fine} or text abstraction/summarization~\cite{xie2022pre}, and have been used to translate between other modalities, such as images and text (called vision-language models~\cite{zhou2022learning,radford2021learning}) in computer vision tasks, or from natural language to code  \cite{wong2023word,jain2022jigsaw}, in which a pretrained task-agnostic language model can be zero-shot and few-shot transferred to perform a certain task~\cite{brown2020language,kojima2022large}. If one accepts the position that KE can be generally described as the process of transforming knowledge in natural language into knowledge in formal language, then it becomes clear that LLMs provide an advance in our ability to perform knowledge engineering tasks.


\section{The use of LLMs in the practice of knowledge engineering: two scenarios}

Given the above discussion, the natural question that arises is: what might be the utility and impact of the use of LLMs for the transformation of natural language into formal language, when applied in the context of the practice of knowledge engineering?

When LLMs emerged as a new technology in the mid-2010s, two views of the relationship between LLMs and knowledge bases (KBs) were put forward. One was the LLM can be a useful component for various processes that are part of a larger knowledge engineering workflow (i.e. "LMs for KBs" \cite{alkhamissi2022review}); the other was that that the LLM is a cognitive artifact that can be treated as a knowledge base in and of itself (i.e., "LMs as KBs" \cite{petroni2019language}). We exploit this dichotomy to formulate a pair of possible future scenarios for the use of LLMs in the practice of KE. One is to use LLMs as a technology for or tool in support of implementing knowledge tasks that have traditionally been build using older technologies such as rule bases and natural language processing (NLP). Another is to use LLMs to remove the need for knowledge engineers to be fluent in a formal language, i.e., by allowing knowledge for a given knowledge task to be expressed in natural language, and then using prompt engineering as the primary paradigm for the implementation of reasoning and learning. We now explore each of these scenarios in turn, and consider the open research problems that they raise.

\subsection{LLMs as components or tools used in knowledge engineering}

We illustrate the first scenario through reference to CommonKADS \cite{schreiber2000knowledge}, a structured methodology that has been used by knowledge engineers since the early 2000's. CommonKADS is the refinement of an approach to providing a disciplined approach to the development of knowledge systems. This approach saw initial development in the nineteen-eighties as a reaction to both the ad-hoc nature of early expert systems development \cite{wielinga1992kads} and to the frequency of failures in the deployment of expert systems in an organizational context \cite{feigenbaum1992personal}. Stemming from early work on making expert systems development understandable and repeatable \cite{hayes1983building}, CommonKADS is distinguished from methodologies more focused on ontology development (e.g., NeON \cite{suarez2011neon}, Kendall and McGuinness's "Ontology 101" framework \cite{kendall2019ontology}, and Presutti's ontology design patterns \cite{presutti2009extreme}) in that it provides practical guidance for specification and implementation of knowledge systems components in a broader sense. It attempts to provide a synoptic guide to the full scope of activities involved in the practice of KE, and show how it relates to the activities of the organization in which that engineering is taking place. As such, in the context of this paper we can use it as a framework to explore for what tasks and in what ways LLMs can be used for KE.

Some tasks identified by CommonKADS as part of the KE process may remain largely unchanged by the use of LLMs. These include knowledge task identification and project organizational design. But others can involve the use of LLMs. LLMs can assist knowledge engineers and/or knowledge providers in the performance of knowledge engineering tasks. They can also be a means for the implementation of modules performing knowledge-intensive tasks. Examples of these uses include the following:

\begin{description}
    \item [Knowledge acquisition and elicitation] LLMs can be used to support knowledge acquisition and elicitation in a given domain of interest. Engineers can create prompts that target specific aspects of the domain, using the responses as a starting point for building the knowledge base. Dialogs between LLMs trained using such prompts and knowledge providers, the subject matter experts, can support the review, validation, and refinement of the acquired knowledge~\cite{bach-etal-2022-promptsource}.
    \item [Knowledge organization] LLMs can be used to organize the acquired knowledge into a coherent structure using natural language, making it easy to understand and update. Prompt engineering can be used to develop a set of prompts that extract formal language using the LLM, e.g., for text to graph generation \cite{guo2020cyclegt} or vice versa~\cite{bosselut2019comet,alivanistos2022prompting}. Moreover, LLMs are used for program synthesis \cite{wong2023word,jain2022jigsaw}, the generation of metadata \cite{keywords2023} or for fusing knowledge graphs \cite{kgpromptfusion2023}.
    \item [Data augmentation] LLMs can be used to generate synthetic training data to aid in testing the knowledge system by evaluating its performance on instances of the specific task~\cite{yoo2021gpt3mix}. 
    \item [Testing and refinement] Feedback from subject matter experts and users can be used to prompt an LLM to refine the natural language knowledge base and improve the system's accuracy and efficiency through self-correction of prompts and tuning of the LLM model settings as needed to optimize the system's performance \cite{white2023chatgpt}. 
    \item [Maintenance] LLMs can be used to monitor new information and trends, and to then propose new prompts integrating those updates into the knowledge base.
\end{description}

\begin{figure}
    \centering
    \includegraphics[width=0.94\textwidth, trim={0 35 0 30},clip]{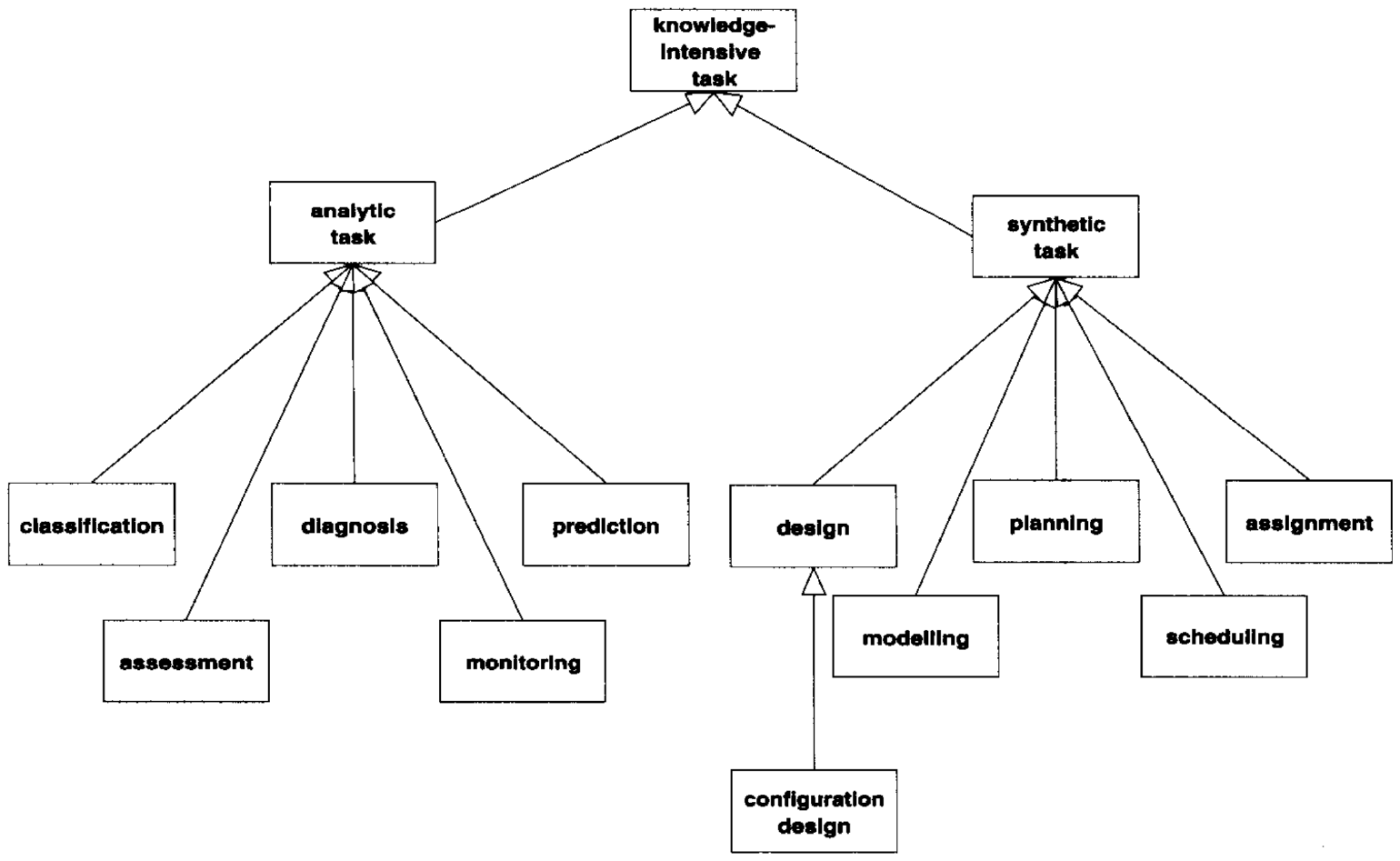}
    \caption{Hierarchy of knowledge-intensive task types from CommonKADS (\cite{schreiber2000knowledge}, p.125)}
    \label{fig:ktasks}
\end{figure}

Consider the CommonKADS knowledge task hierarchy shown in Figure \ref{fig:ktasks}. Synthetic knowledge-intensive tasks, e.g. design or configuration, are amenable to generative approaches \cite{weisz2023toward}; analytic knowledge-intensive tasks can involve LLM components within a hybrid neuro-symbolic knowledge system.

A shortcoming of using CommonKADS for our purposes, however, is that it predates the widespread use of machine learning and statistical  natural language processing in KE. A number of architectural approaches have since been developed that extend the CommonKADS concepts of a knowledge-intensive task type hierarchy and knowledge module templates. These include modeling the fine-grained data flows and workflows associated with knowledge systems that combine components that ingest, clean, transform, aggregate and generate data, as well as generate and apply models built using machine learning \cite{van2019boxology,breit2023combining,daga2023data,ekaputra2023describing,van2021modular}. These architectures are put forward as providing a general framework for composing heterogeneous tools for knowledge representation and inference into a single integrated hybrid neuro-symbolic system. The design pattern notations put forward in recent work \cite{van2019boxology, van2021modular, ekaputra2023describing} treat data, models, and symbolic representations as the inputs and outputs of components composed into a variety of knowledge system design patterns. Generalizing these into natural language and formal language inputs and outputs can provide a simple way to extend these design notations to accommodate both LLMs as well as a richer set of knowledge representations.

\subsection{Knowledge engineering as prompt engineering}

Given that LLMs enable knowledge modeling in natural language, it is conceivable that the programming of knowledge modules could take place entirely in natural language. Consider that prompt programming is "finding the most appropriate prompt to allow an LLM to solve a task" \cite{liu2023pre}. One can through this lens view knowledge engineering as the crafting of dialogues in which a subject matter expert (SME) arrives at a conclusion by considering the preceding context and argumentation \cite{reynolds2021prompt,weisz2023toward,shanahan2022talking,mahowald2023dissociating}. This framing of knowledge engineering as prompt engineering is the second scenario we wish to explore.

From the perspective of the CommonKADS knowledge-intensive task type hierarchy, this would involve a redefinition of the types and hierarchy to use LLMs and prompt programming design patterns, e.g. as described in \cite{liu2023pre}. Several aspects of this redefinition could include:
\begin{description}
    \item [Natural language inference] LLMs can be used to build natural language inference engines that use the organized knowledge to perform the specific task by taking input queries and generate output using prompt engineering to guide the LLM towards generating accurate inferences, e.g. using zero- or few-shot chain-of-thought design patterns. The benefit here is that the gap between the knowledge engineer, knowledge provider (the subject matter expert) and the user is smaller since a translation to a formal language (the language of the engineer) is no longer required. 
    \item [Knowledge-intensive task execution through human/machine dialog] LLMs can be used to a conversational interface that allows users to interact with the knowledge system and receive task-specific support.
    \item [Testing and refinement through human/machine dialog] Feedback from subject matter experts and users can be used to prompt an LLM to refine the natural language knowledge base and improve the system's accuracy and efficiency through self-correction of prompts and tuning of the LLM model settings as needed to optimize the system's performance. 
\end{description}

One possible benefit of this approach would be that the barrier to adoption of knowledge engineering as a practice could be lowered significantly. Knowledge elicitation could be conducted entirely within natural language, meaning that subject matter experts without training in formal knowledge representations could perform these tasks directly. However, this approach assumes that predictable inference \cite{van2021modular} using natural language is satisfactory. 
The propensity of current LLMs to "hallucinate", i.e., to confabulate facts, is an obstacle to the realization of this idea \cite{ji2023survey}. Multiple efforts have been devoted to the creation of prompt programming patterns that address this issue, ranging from chain-of-thought approaches \cite{wei2022chain} to retrieval-assisted generation, i.e. the augmentation of LLMs with authoritative document indexes and stores \cite{schick2023toolformer,mialon2023augmented}. Recent work \cite{pan2023unifying} has described ways in which knowledge graphs as a formal language can be integrated with natural language and LLM-based language processing and reasoning to provide knowledge systems architectures that directly address this issue. \cite{yang2023logical} surveys work in this direction.

\section{Open research questions}

Using the scenarios outlined above, we can identify a number of open research questions to be addressed to realize either or both of these two possible approaches to the use of LLMs in knowledge engineering. These questions touch on three general areas: the impact of LLMs on the methodologies used to build knowledge systems, on the architectural design of knowledge systems incorporating and/or based on LLMs, and on the evaluation of such systems. For each of these open questions, we provide a link back to the biodiversity scenario discussed in Section \ref{sec:bio} denoted by a \faLeaf.

\subsection{Methodology}

\subsubsection{How can knowledge engineering methodologies best be adapted to use LLMs?}

How can we harmoniously meld the considerable body of work on knowledge engineering methodologies \cite{kendall2019ontology,gangemi2009ontology,presutti2009extreme,suarez2011neon,schreiber2008principles,schreiber2008knowledge,staab2010handbook} with the new capabilities presented by LLMs? 

Schreiber's conceptualization of knowledge engineering as the construction of different aspect models of human knowledge \cite{schreiber2000knowledge}, as discussed above, offers a framework for further elaboration. The distinctive characteristics of LLMs, coupled with prompt engineering, present unique challenges and opportunities for building agents within a knowledge system, one that is consistent with the CommonKADS approach.

While the role definitions within KE methodologies might mostly remain the same, the skills required for knowledge engineers will need morphing to adapt to the LLM environment. This evolution of roles calls for an extensive investigation into what these new skills might look like, and how they can be cultivated. Additionally, the adaptability of the various knowledge-intensive task type hierarchies described by CommonKADS and its descendants in the literature on hybrid neuro-symbolic systems (e.g., as described in \cite{breit2023combining}) to accommodate LLMs is another fertile area for exploration.

LLM-based applications, likened to synthetic tasks within these knowledge engineering frameworks, raise compelling research questions regarding accuracy and the prevention of hallucinations. LLM-based applications have a lower bar to reach with respect to notions of accuracy and avoidance of hallucinations, but still must provide useful and reliable guidance to users and practitioners.

\faLeaf~Connecting back to the biodiversity domain, answering these questions would provide guidance on the appropriate methodology to adopt when developing a new specimen curation and collection knowledge management system that needs to deal with multimodal assets like handwritten text or images.

\subsubsection{How do principles of content and data management apply to prompt engineering?}

Applying content and/or data management principles to collections of prompts and prompt templates, integral to work with LLMs, is an area ripe for exploration. Properly managing these resources could improve efficiency and guide the development of improved methodologies in knowledge engineering. This calls for a rigorous investigation of current data management practices, their applicability to LLMs, and potential areas of refinement.
Ensuring the reproducibility of LLM engineering from a FAIR data standpoint \cite{Wilkinson2016} is a crucial yet complex challenge. Developing and validating practices and protocols that facilitate easy tracing and reproduction of LLM-based processes and outputs is central to this endeavour. 

\faLeaf~Addressing this challenge will aid researchers in applying LLM engineering in a FAIR way. Doing so is critical for biodiversity research and science in general where precision, reproducibility and provenance are key for knowledge discovery and research integrity.

\subsubsection{What are the cognitive norms that govern the conduct of KE?}

A crucial area of inquiry involves the identification and understanding of \textit{cognitive norms}, as described by Menary \cite{menary2007writing}, that govern the practice of knowledge engineering. Cognitive norms are established within a human community of practice as a way of governing the acceptable use of "external representational vehicles to complete a cognitive task" \cite{menary2010dimensions}. As the consumer adoption of LLM technology has progressed, we see a great deal of controversy about when and how it is appropriate to use, e.g. in the context of education or the authoring of research publications. Understanding how these norms shape the use of LLMs in this context is an under-explored field of study. By unravelling the interplay between these cognitive norms and LLM usage, we can gain valuable insights into the dynamics of knowledge engineering practices and possibly foster more effective and responsible uses of LLMs.

\faLeaf~In the biodiversity sciences, this means understanding the cognitive norms specific to the domain, to understand how LLMs can be used in a way that respects the domain's practices and standards.   

\subsubsection{How do LLMs impact the labor economics of KE?}

A related but distinct question pertains to the impact of LLMs on the economic costs associated with knowledge engineering. The introduction and application of LLMs in this field may significantly alter the economic landscape, either by driving costs down through automation and efficiency or by introducing new costs tied to system development, maintenance, and oversight. Thoroughly exploring these economic implications can shed light on the broader effects of integrating LLMs into knowledge engineering.

The realm of labor economics as it pertains to hybrid or \textit{centaur} systems \cite{akata2020research}, is another area ripe for investigation. Understanding how the deployment of these systems influences labor distribution, skill requirements, and job roles could provide valuable input into the planning and implementation of such technologies. Additionally, it could reveal the potential societal and economic impacts of this technological evolution.

\faLeaf~Developments for LLM-based KE can help mitigate labour of knowledge experts in the biodiversity sciences, for instance by the development of more efficient KE workflows for the digitization of museum specimens or manuscripts.  

\subsection{Architecture}

\subsubsection{How can hybrid neuro-symbolic architectural models incorporate LLMs?}

Design patterns for hybrid neuro-symbolic systems, as described in \cite{van2019boxology}, offer a structured approach to comprehend the flow of data within a knowledge system. Adapting this model to account for the differences between natural and formal language could significantly enhance our ability to trace and manage data within knowledge systems.
A salient research question emerging from this scenario pertains to the actual process of integrating LLMs into knowledge engineering data processing flows \cite{daga2023data}. Understanding the nuances of this process will involve a deep examination of the shifts in methodologies, practices, and the potential re-evaluations of existing knowledge engineering paradigms. The perspective of KE enabled by LLMs as focused on the transformation of natural language into formal language provides insights that can be used to improve the motivation for hybrid neuro-symbolic systems; e.g., \cite{breit2023combining} references \cite{booch2021thinking} in using dual process theories of reasoning (i.e. the "System 1/System 2" model described in \cite{kahneman2011thinking}) as a motivation for hybridization in knowledge systems, but more recent analyses \cite{novaes2012formal, mercier2017enigma} cast doubt on the validity of such models, and point to more nuanced perspectives that provide a better grounding for the benefits of hybridization.

\faLeaf~Addressing these questions would shed light on tasks for which hybridization using LLMs would prove favourable, e.g., image classification of species.

\subsubsection{How can prompt engineering patterns support reasoning in natural language?}

One fundamental question that arises is how prompt engineering patterns can be utilized to facilitate reasoning in natural language. Exploring this topic involves understanding the mechanics of these patterns and their implications on natural language processing capabilities of LLMs. This line of research could open new possibilities for enhancing the functionality and efficiency of these models.

A related inquiry concerns the structure, controllability, and repeatability of reasoning facilitated by LLMs. Examining ways to create structured, manageable, and reproducible reasoning processes within these models could significantly advance our capacity to handle complex knowledge engineering tasks and improve the reliability of LLMs.

The interaction of LLMs and approaches to reasoning based on probabilistic formalisms is also an underexplored area of research. A particularly evocative effort in this area is that described in \cite{wong2023word}, which describes the use of LLMs to transform natural language into programs in a probabilistic programming language, which can then be executed to support reasoning in a particular problem domain. We note that this work provides an excellent example of the knowledge engineering as the transformation of natural language into formal language perspective and of the impact of LLMs in advancing that perspective. Investigating how to automatically generate and assess other nuanced forms of knowledge within LLMs could lead to a more refined understanding of these models and their capabilities.

\faLeaf~Given that biodiversity knowledge is often best represented in a variety of modalities each with their own data structures and characteristics, research may explore how LLMs can act as natural language interfaces to such multimodal knowledge bases.   

\subsubsection{How can we manage bias, trust and control in LLMs using knowledge graphs?}

Trust, control, and bias in LLMs, especially when these models leverage knowledge graphs, are critical areas to explore. Understanding how to detect, measure, and mitigate bias, as well as establish trust and exert control in these models, is an essential aspect of ensuring ethical and responsible use of LLMs.
Furthermore, investigating methods to update facts in LLMs serving as knowledge graphs is a crucial area of research. Developing strategies for efficient and reliable fact updating could enhance the accuracy and usefulness of these models.

Another key question involves understanding how we can add provenance to statements produced by LLMs. This line of research could prove vital in tracking the origin of information within these models, thus enhancing their reliability and usability. It opens the door to more robust auditing and validation practices in the use of LLMs.

\faLeaf~Addressing this challenge can help biodiversity researchers detect and mitigate biases, as use of LLMs might further exacerbate knowledge gaps, e.g., groups of individuals omitted from historical narratives in archival collections. Moreover, novel update mechanisms can aid researchers to reliably update facts or changing knowledge structures learned by LLMs, for instance when domain knowledge evolves.   

\subsubsection{Is extrinsic explanation sufficient?}

A significant area of interest pertains to how we can effectively address the explainability of answers generated using LLMs \cite{dovsilovic2018explainable}. This exploration requires a deep dive into the functioning of LLMs and the mechanisms that govern their responses to prompts. Developing a thorough understanding of these processes can aid in creating transparency and trust in LLMs, as well as fostering their effective use.

The need for explanation in LLMs also leads to the question of whether extrinsic explanation is sufficient for the purposes of justifying a knowledge system's reasoning, as argued in general for the intelligibility of knowledge systems by Cappelen and Devers \cite{cappelen2021making}, or if intrinsic explainability is a necessary requirement \cite{DBLP:journals/corr/abs-2202-01875}. This question calls for a thoughtful exploration of the value and limitations of both extrinsic and intrinsic explanation methodologies, and their implications for the understanding and usage of LLMs.
An exciting research avenue arises from the work of Tiddi \cite{tiddi2022knowledge}, concerning explainability with formal languages. The exploration of this topic could reveal significant insights into how we can leverage formal languages to enhance the explainability of LLMs. This could pave the way for new methods to increase transparency and intelligibility in these models.

\faLeaf~ In the sciences in general, answering these questions would aid explainability of LLM-generated answers via curated facts, increasing transparency and trust.  

\subsubsection{How can LLMs support the engineering of hybrid human/machine knowledge systems?}

Another topic of interest involves exploring the potential of hybrid systems that combine human cognition with machine capabilities within a dialogical framework \cite{mercier2017enigma, novaes2020dialogical}. As an exciting example of the possibilities for new approaches to human/machine collaboration in this vein, we point to the recent results reported by \cite{park2023generative} on the creation of conversational agents that simulate goal-directed human conversation and collaboration on tasks. One can imagine coupling LLM-based agents with human interlocutors working collaboratively in this manner on specific knowledge-intensive tasks. Understanding how to develop these types of systems, and what their implications might be for the practice of knowledge engineering presents a fertile research line. It requires the careful analysis of human-machine interaction, the study of system design principles, and the investigation of their potential impact.

\faLeaf~Research in this avenue can help mitigate the workload of the knowledge expert, for instance in the elicitation of domain knowledge, or crowdsourcing of annotations from unstructured sources such as herbaria or manuscripts. 

\subsection{Evaluation}

\subsubsection{How do we evaluate knowledge systems with LLM components?}

The first point of interest involves the evaluation of knowledge-based systems, with a focus beyond just logic. This area calls for innovative methodologies to assess the system's capacity to manage and utilize knowledge efficiently, going beyond traditional logical evaluations.
This topic of evaluation naturally extends to the question of how we evaluate ontologies and design patterns within knowledge engineering. Evaluating these aspects would require a deep dive into the structures and mechanisms underpinning these elements, potentially leading to the development of refined evaluation metrics and methodologies.

Interestingly, the long-standing paradigm of machine learning evaluation, relying on benchmarking against a standard train/test dataset, seems to falter in the era of LLMs \cite{chang2023survey}. This presents an intriguing challenge for researchers and engineers alike. It is quite possible that  traditional methods may need to be significantly buttressed by methodologies and supporting tools for the direct human evaluation of knowledge system performance. This has implications concerning the cost and speed of evaluation processes, encouraging the rethink of current approaches to perhaps develop new strategies that balance accuracy, cost-effectiveness, and timeliness. Reimagining evaluation methodologies in this new context could provide transformative insights into how we can gain confidence in the reliability engineering of knowledge systems that use LLMs.

\faLeaf~Developments in this direction may aid biodiversity researchers to get a better understanding of the real-world efficacy of employing knowledge-based systems with LLM components in their institutions. One can think of improving access to collections, knowledge discovery, or accuracy in describing institutional knowledge. 

\subsubsection{What is the relationship between evaluation and explainability?}

Lastly, there is an inherent dependency of evaluation on effective solutions for explainability within knowledge systems. Understanding this relationship could help in the creation of more comprehensive evaluation models that take into account not only the performance of a system but also its explainability.

\section{Summary}

In this paper, we have advocated for a reconsideration of the practice and methodology of knowledge engineering in light of the emergence of LLMs. We argued that LLMs allow naturally-occurring and humanly-evolved means of conveying knowledge to be brought to bear in the automation of knowledge tasks. We described how this can enhance the engineering of hybrid neuro-symbolic knowledge systems, and how this can make knowledge engineering possible by people who do not necessarily have the experience of recasting natural language into formal, structured representation languages. Both of these possibilities will involve addressing a broad range of open questions, which we have attempted to outline above. Given the rapid pace of the development of this area of research, it is our earnest hope that the coming months and years will yield results shedding light on these questions.

\subparagraph*{Acknowledgements}
This work was partially supported by the EU’s Horizon Europe research and innovation programme within the ENEXA project (grant Agreement no. 101070305).


\newpage 
\bibliographystyle{plainurl}
\bibliography{tgdkbib}

\end{document}